# Efficient low-thrust trajectory data generation based on generative adversarial network

Ruida Xie [1] and Andrew G. Dempster [2]
*University of New South Wales, Sydney, New South Wales, 2052, Australia*

**Deep learning-based techniques have been introduced into the field of trajectory optimization in recent years. Deep Neural Networks (DNNs) are trained and used as the surrogates of conventional optimization process. They can provide low thrust (LT) transfer cost estimation and enable more complex preliminary mission designs. However, it is a challenge to efficiently obtain the required amount of trajectory data for training. A Generative Adversarial Network (GAN) is adapted to generate the feasible LT trajectory data efficiently. The GAN consists of a generator and a discriminator, both of which are deep networks. The generator generates fake LT transfer features using random noise as input, while the discriminator distinguishes the generator's fake LT transfer features from real LT transfer features. The GAN is trained until the generator generates fake LT transfers that the discriminator cannot identify. This indicates the generator generates low thrust transfer features that have the same distribution as the real transfer features. The generated low thrust transfer data have a high convergence rate, and they can be used to efficiently produce training data for deep learning models. The proposed approach is validated by generating feasible LT transfers in a Near-Earth Asteroid (NEA) mission scenario. The convergence rate of GAN-generated samples is 84.3%.**

## I. Introduction

An increasing number of studies use deep learning techniques for trajectory optimization in landing [1-3] or interplanetary multi-target mission scenarios [4-6]. Deep learning models are trained and used as the surrogate of the conventional optimization. The major motivation of utilizing such techniques is that they provide estimates of solutions with an acceptable level of accuracy using a much shorter solving time, which enables the trajectory design in more complex contexts, such as multi-target interplanetary transfers [6, 7], real-time guidance [3, 4] and robust trajectory planning in severe uncertainties and disturbances [8]. However, training high performance deep learning models requires large databases. Hence, efficiently generating the trajectory database is a common problem in this emerging area.

This study focuses on the efficient trajectory database generation for low thrust (LT) trajectory optimization for near-Earth asteroid (NEAs) missions. In this context, Deep Neural Networks regressors are often trained to predict low thrust transfer cost, such as propellant consumption [5, 6, 9] and transfer time [10], without performing conventional direct or indirect optimization. The DNNs can process thousands of transfers in seconds, so that complex missions, such as multi-target rendezvous [6, 11-14], can be designed efficiently.

During training data generation, the obstacle is low convergence rate. The conventional data generation method is illustrated in Fig. 1. Two asteroids are first randomly selected. Transfer parameters, such as initial mass and time of flight are then initialized. The low thrust transfer is then optimized based on these parameters in either a direct or indirect way [15]. The convergence and divergence of the optimization corresponds to a feasible and infeasible low thrust transfer, respectively. Regardless of the optimization method selected, the feasibility of a low thrust transfer depends on the ephemerides of asteroids and transfer parameters. In our testing scenario, converging samples account for 7.16% of all generated samples.

---

[1] PhD candidate, Australian Centre for Space Engineering Research (ACSER), School of Electrical Engineering and Telecommunications, ruida.xie@student.unsw.edu.au.
[2] Professor, Australian Centre for Space Engineering Research (ACSER), School of Electrical Engineering and Telecommunications, a.dempster@unsw.edu.au

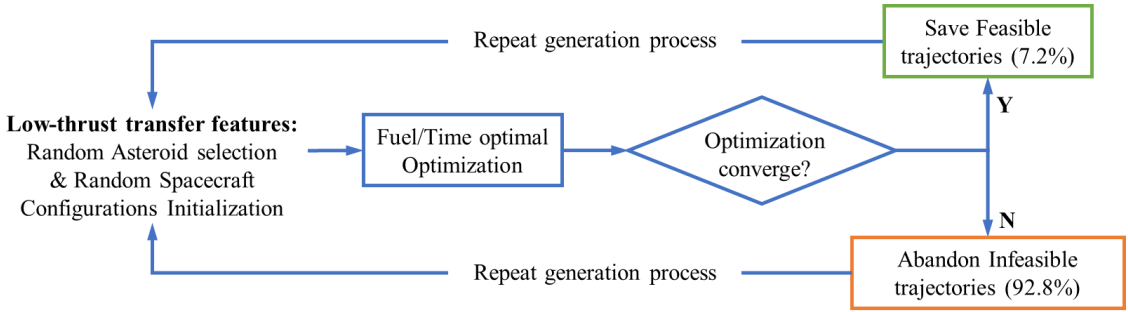

**Fig. 1 Conventional workflow for trajectory data generation.**

Previous studies have recognized this problem, and several measures have been taken to improve the data generation efficiency. In Ref. [6, 10], the conventional process is improved by limiting the difference of orbital elements between two targets, such as inclinations, semi-major axes and true longitudes, etc. By using this method, the chosen asteroid pairs have similar orbits and higher probabilities of convergence. In Ref. [14, 16], phase indicators are constructed to quantify the similarity between two targets' orbits. However, these approaches sacrifice the database integrity: some excluded samples may be feasible transfers.

To tackle this issue, we propose a new approach to directly generate feasible transfer pairs via use of a generative adversarial network (GAN). Generally, the GAN can be seen as the surrogate of the conventional LT feature generation process. The advantage of the GAN is that it learns how to generate only feasible LT transfers that will converge in the optimization process. The GAN uses noise vectors as inputs and does not require real asteroid data, but it can convert the noise vectors into LT transfer features that have the same distribution as real ones. Statistically, a GAN trained in this work improves the convergence rate from 7.16% to 82%.

The remaining sections are as follows: Section II introduces the training set. Section III introduces the structure of GAN. Section IV evaluates the GAN-produced database. Section V concludes the paper.

## II. Real transfer data generation

A GAN needs real LT transfers to learn the distribution. The original LT transfer data are generated following the process illustrated in Fig. 1, repeatedly applying direct optimization [17] on randomly initialized transfers between selected near-Earth asteroids (NEAs).

The ephemerides of NEAs are obtained from the JPL HORIZONS system. To increase the convergence rate, the maximum absolute differences between two asteroids' semi-major axes, inclinations and true longitudes are limited to $0.3\ AU$, $3°$ and $30°$, respectively.

The spacecraft initial mass is assumed to be any value in the range $[1,000,\ 3,000]kg$. The spacecraft dry mass is 1,000 kg. Any optimised final mass equal to or below this value means the spacecraft has exhausted the propellant, thus the LT transfer is considered to be infeasible. The spacecraft is assumed to be equipped with a Xenon Thruster having maximum thrust force of $0.236\ N$ and a specific impulse of $4,190\ s$, which provides an effective exhaust velocity of $41.09\ km/s$.

An impulsive transfer grid search is then performed to provide the $\Delta t_{impls}$, the time of flight for the minimum impulsive delta-V transfer. Considering that the time of flight for LT transfers is generally longer than for impulsive transfers, the LT transfer time of flight $\Delta t_{LT}$ is initialised as $1.2 \cdot \Delta t_{impls} \leq \Delta t_{LT} \leq \min\{2 \cdot \Delta t_{impls}, 4 \cdot 365\}$. To avoid local minima, the direct optimisation is performed 5 times for each feasible transfer, then the solution with the largest final mass is recorded in the database.

The real transfer data along with the generated data will be sent to the discriminator to calculate the loss for the network update. Details of this are in Section III.

## III. Low thrust trajectory generation via GAN

### A. GAN Workflow Overview

A GAN is a type of deep neural network (DNN). The architecture comprises two DNNs, a generator and a discriminator, which work against each other during training. The overall objective of training the GAN is to obtain a generator that only generates feasible low thrust (LT) transfer features, such that the convergence rate of optimization of these transfers can be improved. The training data flow, network updating, and GAN utilization workflow is illustrated using different colored arrows in Fig. 2.

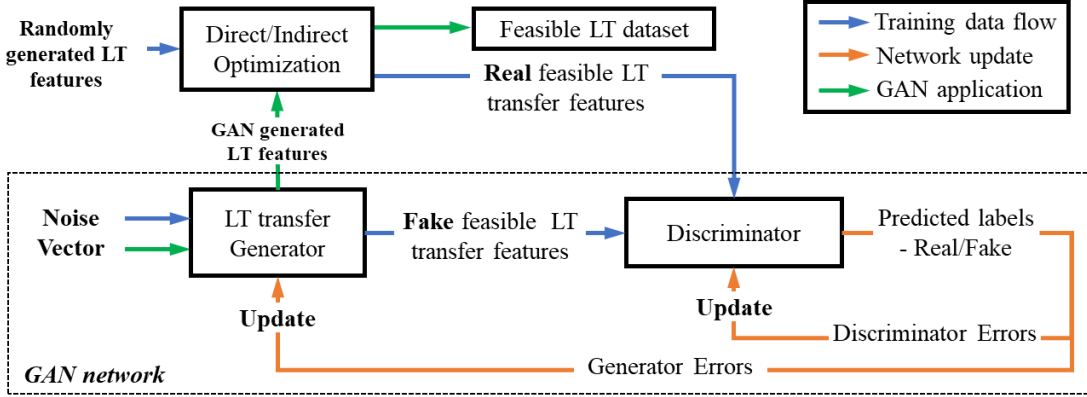

**Fig. 2 GAN network for producing feasible low thrust transfers**

The GAN has two types of data inputs: the random noise fed to the generator and the real low thrust transfer features fed to the discriminator. The generator uses Gaussian noise as input to generate fake feasible LT data instances. Then an equal number of real and fake LT features are fed to the discriminator. The discriminator evaluates the LT features for authenticity and decides whether each instance of data is real or fake from the generator. The losses for the generator and discriminator are then calculated, and the entire network is updated.

The generator and discriminator are trained to work against each other until the generator produces feasible LT transfer features that 'fool' the discriminator.

### B. Network Structure and Hyperparameter selection

The noise is fed to the generator batch by batch. Each batch is a $M \times N_b$ noise matrix, where $M$ is the length of the noise vector, $N_b$ is the batch size. The outputs of the generator are normalised LT transfer features. The LT transfer features are derived from Ref. [5], including spacecraft initial mass, time of flight, Modified Equinoctial Elements (MEEs) of the two targets, the differences between MEEs $\Delta MEE$, the orbital energy difference $\Delta E$ and angular momentum difference $\Delta H$. For the GAN-generated samples, the $MEEs$ do not necessarily correspond to any existing NEAs.

The inputs of the discriminator are the mixed dataset consisting of the output of the generator and the real feasible LT transfer features. Each input index corresponds to a label indicating whether they are the generated 'fake' or real transfer features. The outputs of the discriminator are the predicted labels.

The hyperparameter search ranges are as shown in **Table 1**.

**Table 1 Hyperparameters range for GAN**

| Parameters | Generator | Discriminator |
|---|---|---|
| Layer Numbers $L$ | [5, 30] | [4, 10] |
| Neurons $n_{neurons}$ | [100,300] | [200,400] |
| Drop-out rate $r_{drop}$ | [0,0.8] | [0,0.8] |
| Learning rate $\eta$ | [1e-5,3e-4] | [1e-5,3e-4] |

Both the generator and the discriminator use fully connected hidden layers. The differences are network dimensions, such as the number of layers $L$, number of neurons $n_{neurons}$, and dropout rate $r_{drop}$. These hyperparameters need to be carefully chosen. In this work, a grid search is performed to decide a workable GAN structure. In the preliminary test runs, we found the generator requires a much deeper structure than the discriminator. Thus, the search ranges for the depths of generator and discriminator are different.

**C. Training and evaluation**

Training and selecting appropriate hyperparameters for the GAN is a challenging task. Because the GAN structure is more complex than normal deep networks, inappropriate choice of a single parameter may lead to training divergence and mode collapse. Meanwhile, the generator and the discriminator networks compete against each other during the training, and it is critical to balance the power of the generator and discriminator. If one model learns too fast, then the other one may fail to converge.

Divergence of training refers to the condition where the generator is unable to generate feasible low thrust transfer features. A convergence failure happens when the generator and discriminator do not reach a balance during training, which is caused by discriminator dominance or generator dominance. In this case, discriminator *dominance* is more often the case. This can be identified by observing the test accuracy. A low convergence rate of testing samples corresponds to a convergence failure.

A mode collapse refers to the situation when the GAN produces a small variety of LT features. The cause of this is that the generator is too weak to learn the rich LT feature representation, as it learns to associate similar outputs with multiple different inputs. In the field of image generation, a GAN with mode collapse generates almost identical pictures. In this study, a GAN with mode collapse generates LT features which are subsets of the real ones. A typical mode collapse is illustrated in Fig. 6 and Fig. 7 in Section IV.C.

Compared to training divergence, mode collapse cannot be identified solely based on accuracy, because a mode collapse GAN usually generates LT transfers having a high convergence rate.

To identify GAN divergence and mode collapse, the training process is monitored using the generator score $s_{gen}$ and discriminator score $s_{dis}$. The discriminator always outputs the probability of the LT features being 'real'. The $s_{gen}$ is the mean of the probabilities for the generated LT features, while the $s_{dis}$ is the mean of the probabilities of the input LT features belonging to the correct classes. The $s_{gen}$ and $s_{dis}$ are given by:

$$\begin{cases} s_{gen} = \frac{1}{N_b} \sum_{1}^{N_b} \hat{y}_{gen} \\ s_{dis} = \frac{1}{2} \cdot \frac{1}{N_b} \sum_{1}^{N_b} \hat{y}_{real} + \frac{1}{2} \cdot \frac{1}{N_b} \sum_{1}^{N_b} (1 - \hat{y}_{gen}) \end{cases} \quad (1)$$

where the $N_b$ is the batch size, $\hat{y}_{gen}$ and $\hat{y}_{real}$ are the probabilities of the generated LT features and the real LT features to be classified as real. Ideally, both scores would be 0.5, which means the discriminator cannot tell the difference between the generated LT data and the real data. However, this is not a necessary condition to obtain a successful GAN in this work. A successful GAN trained in this work has a generator score that converges to 0.3 - 0.4 and a discriminator score that converges to 0.6 - 0.7.

The score is an important tool to be used to create early stopping criteria during the GAN training, to avoid the convergence failure and model collapse.

As we need to monitor the training process of the GAN to identify mode collapse, advanced hyperparameter optimization techniques, such as Bayesian Optimization, cannot be used very much in this case. Instead, a grid search method is used, and the final parameter settings are determined by experience.

Because we do not have any prior knowledge of applying GAN in this field, deciding on an appropriate GAN structure using grid search requires us to explore a large search space. Meanwhile, training the GAN is time-consuming. The training of DNN-based LT cost approximation models usually converges within 50 epochs [1, 5, 6], while the GAN in this study requires hundreds of epochs before converging. Therefore, the grid search computation was performed on: 1) UNSW High Performance Computing (HPC) cluster Katana; 2) HPC cluster Gadi from National Computational Infrastructure (NCI) supported by UNSW High Performance Computing Resource Allocation Scheme (Project: im32).

## IV. Experiment

### A. Real low thrust transfer features

This section presents the statistical results for real LT transfer features. The summary of the 30,000 feasible LT transfers is given in Table 2.

**Table 2. Overview of the feasible LT database**

|        | $\Delta t_{LT}$ | $m_i$  | $\Delta p$  | $\Delta f$  | $\Delta g$  | $\Delta h$  | $\Delta k$  | $\Delta L$ |
|--------|-----------------|--------|-------------|-------------|-------------|-------------|-------------|------------|
| Min    | 126             | 1053   | -3.30e-01   | -3.82e-01   | -3.97e-01   | -2.85e-01   | -2.86e-01   | -6.28      |
| Max    | 1460            | 3000   | 3.30e-01    | 3.82e-01    | 3.97e-01    | 2.77e-01    | 2.87e-01    | 6.25       |
| Mean   | 875.9           | 1830.5 | -4.43e-03   | 5.98e-04    | -3.11e-04   | 8.06e-05    | 1.23e-04    | -5.63e-03  |
| Median | 883             | 1732   | -4.29e-03   | 7.69e-04    | -3.69e-04   | 5.01e-05    | 5.45e-05    | 3.37e-03   |

In Table 2, $\Delta t_{LT}$ is the low thrust transfer time, $m_i$ is the initial mass, the difference between two bodies' Modified Equinoctial Elements (MEEs) are given by $\Delta p, \Delta f, \Delta g, \Delta h, \Delta k, \Delta L$. The scatter plot of the real feasible low thrust features is given in Fig. 3. The features are scaled to the range of -1 ~ +1. The diagonal plots show the histograms for corresponding features. It is notable that there is a peak of the true longitude ($\Delta L$) in the right bottom subplot. This occurs because we added a limitation on differences between two asteroids' longitudes as introduced in Section II. The performance of the GAN can be evaluated by comparing the distributions of GAN generated data and real data in the matrix plot.

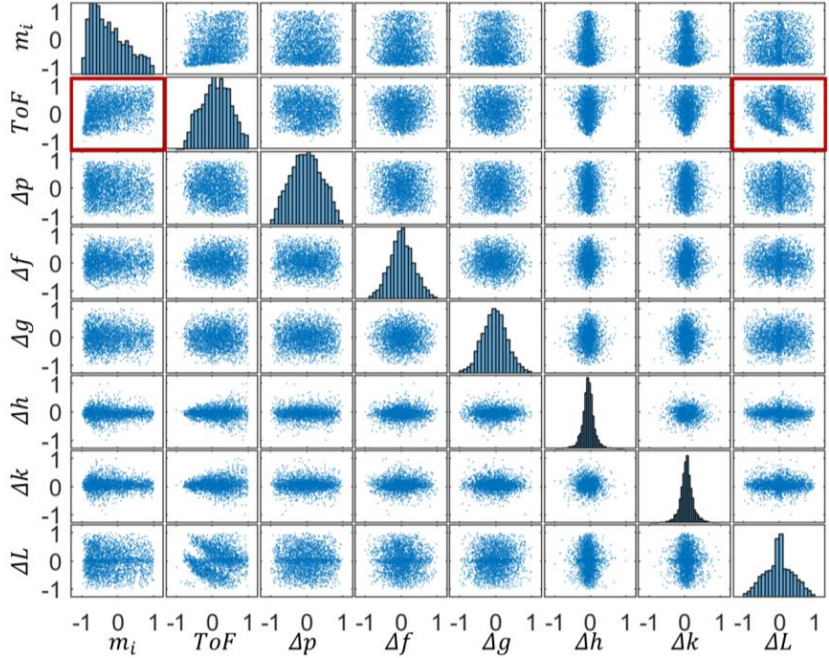

**Fig. 3 Real LT transfer feature distribution. Each diagonal subplot shows the histogram of the corresponding feature. Each off-diagonal subplot is a scatter plot of a column of a LT feature against another column of a LT feature.**

### B. Training process of the GAN

*1. Best observed GAN structure*

The GAN hyperparameters were searched using the High-Performance Computing cluster. The computation consumed 4,583 core-hours of computational resources, where 3,577 core-hours were consumed for training and 1006 hours for performance validation. The search step for the layer number, number of neurons drop-out rate and learning rates are 2, 50, 0.1 and 0.00005, respectively. The best observed GAN configurations are shown in Table 3. The test convergence rate is 84.3%.

Table 3. Best observed GAN parameters

| Parameters | Generator | Discriminator |
|---|---|---|
| Layer Numbers $L$ | 20 | 6 |
| Neurons $n_{neurons}$ | 250 | 300 |
| Drop-out rate $r_{drop}$ | 0.5 | 0.4 |
| Learning rate $\eta$ | 0.00015 | 0.0002 |

From the table we see that the generator has much deeper but thinner structure than the discriminator. The dropout out rate of the generator is 10% higher than the discriminator, and the learning rate is 25% lower. This can be explained as the deeper DNN of generator requiring a lower learning rate to avoid divergence.

The GAN optimization is based on a grid search. Therefore, the best observed model is not necessarily optimal. Besides the listed parameters, there are many other optimizable parameters, such as batch size $N_b$, flip factor and optimiser related parameters. These parameters are less important to the convergence of training but tuning these parameters may further improve the performance.

*2. Training of a successful GAN*

The training process of a successful GAN is shown in Fig. 4. The GAN is trained for 40,000 iterations (800 epochs). The epoch indicates the number of passes of the entire training dataset the GAN has completed.

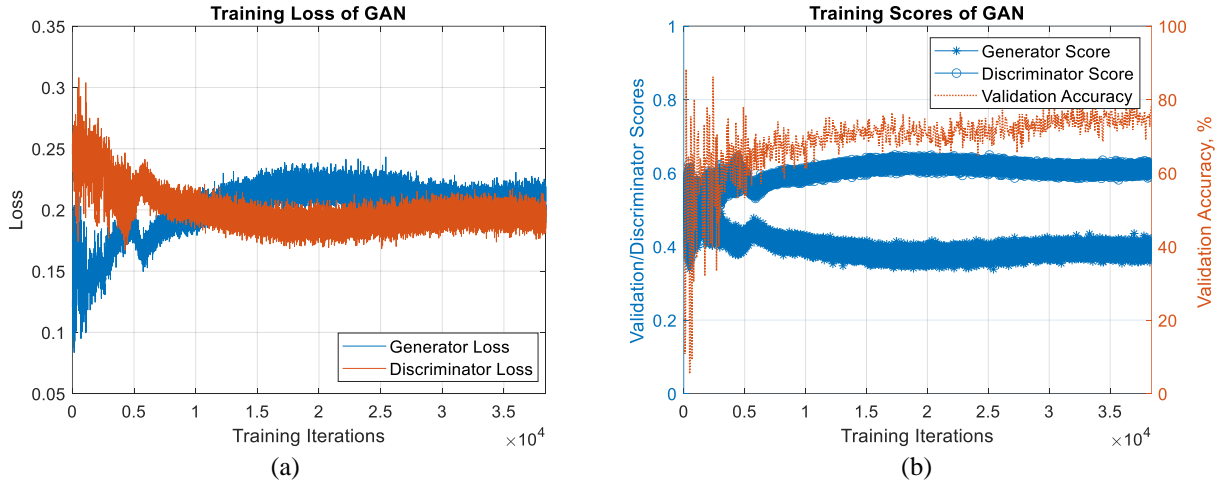

**Fig. 4** Training process of the best observed GAN. (a) Training loss of the GAN. (b) Scores and validation accuracy plot.

Fig. 4 (a) shows the training loss of the generator and discriminator. It can be seen that the generator loss and discriminator loss both converge to approximately 0.2 after 30,000 iterations of training Fig. 4 (b) shows the scores of the generator and discriminator, and the validation accuracy. Scores are designed to evaluate the power of the generator and discriminator. As shown in the plot, the discriminator score and generator score converge to 0.6 and 0.4 respectively.

The dashed red line in Fig. 4 (b) is the validation accuracy. The validation frequency is set as:

$$f_{valid} = \frac{N_{train}}{N_b} \qquad (2)$$

where $N_{train}$ is the number of the training samples and $N_b$ is the batch size, such that the GAN is validated once in each epoch. To enable a fast validation during training, the validation accuracy is estimated by using a Deep Neural

Network Classifier proposed in Ref. [18]. Both the discriminator and the classifier are binary classification models. The difference is that the classifier is originally designed to predict the feasibility of low thrust transfers that are randomly initiated like the process introduced in Section II, while the discriminator of the GAN is to distinguish real transfer features from fake transfer features produced by the GAN generator. We found that the actual convergence rate is slightly higher than the value that is predicted by the DNN-classifier. This indicates that the generator creates samples that are the outliers for the DNN-classifier but are in fact feasible.

The DNN-classifier is mainly used as a qualitative rather than a quantitative index. It is used to monitor the stability of the GAN during training. In Fig. 4 (b), the validation accuracy strongly fluctuates in the first 100 epochs, but it becomes stable after 200 epochs. Although the loss and score are still changing after 200 epochs, the validation curve shows the model tends to converge.

The generated data is plotted in the same style as Fig. 3. By comparing Fig. 3 and Fig. 5, we find that GAN produced sample points have a similar distribution to the real data points. Even for the subplots marked by the red rectangle, where the distributions have irregular shapes, the GAN still performs well.

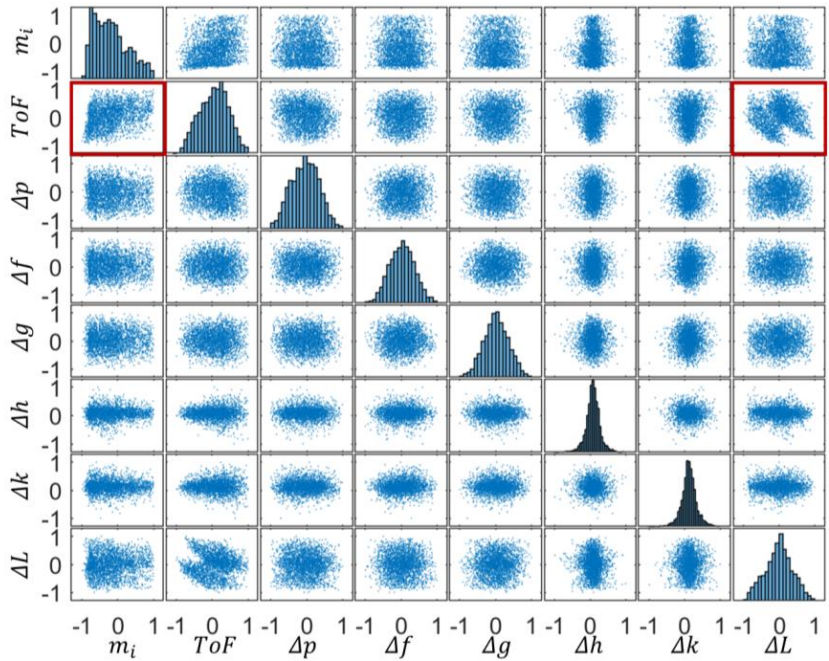

**Fig. 5 Distribution of generated fake feasible LT samples from the best observed GAN**

### C. Identifying mode collapsed GAN

The training of the GAN is very sensitive to some parameters listed in Table 3. The structure of a collapsed GAN is given in **Table 4**. The main factors that influence the stability of the GAN is the change of dropout rate and learn rate. The dropout rate of the discriminator increases by 0.1. The learn rate for the generator increases by $5 \times 10^{-5}$, and the discriminator learn rate drops by $2.5 \times 10^{-5}$.

**Table 4. Mode collapsed GAN parameters**

| Parameters | Generator | Discriminator |
|---|---|---|
| Layer Numbers $L$ | 20 | 6 |
| Neurons $n_{neurons}$ | 300 | 250 |
| Drop-out rate $r_{drop}$ | 0.5 | 0.5 |
| Learning rate $\eta$ | 0.0002 | 0.000175 |

The training process of the collapsed GAN is shown in Fig. 6. The loss of the generator and discriminator quickly converge to around 0.13 and 0.25, respectively. Meanwhile, the scores of both two DNNs reach 0.5. The test convergence rate is 97%. This looks like a perfectly trained GAN. However, the fluctuated validation accuracy shows the GAN is not in a stable state.

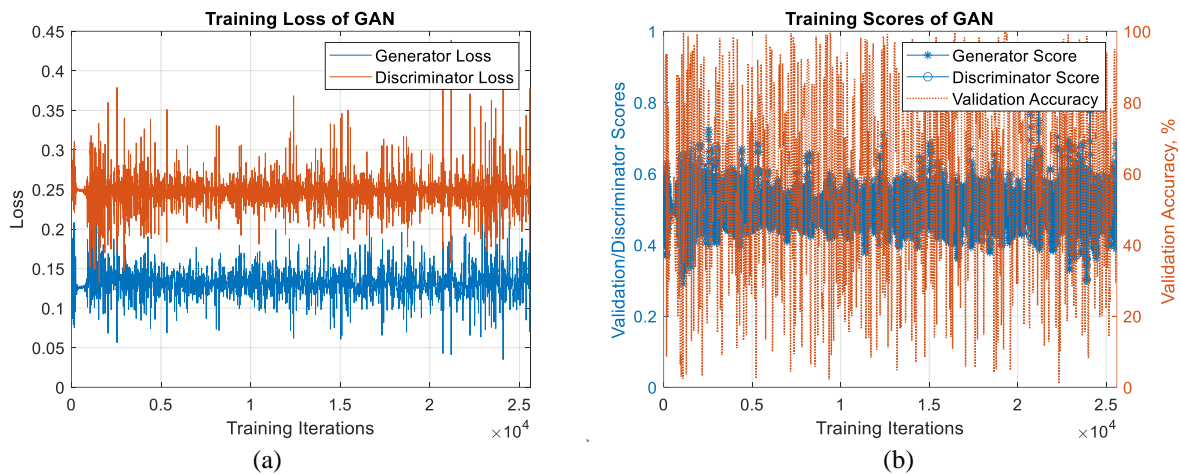

**Fig. 6  Training process of a collapsed GAN. (a) Training loss of the GAN. (b) Scores and validation accuracy plot.**

The distribution of the generated LT features of a mode collapsed GAN is shown in Fig. 7. The orange points represent the LT features generated from the mode collapsed GAN, the blue sample points represent the real LT features. The diagonal subplots show the distribution comparison of the two kinds of samples.

It can be seen from the matrix plot that the collapsed GAN finds a way to fool the discriminator - samples from collapsed GAN are only distributed on a partial area of the real sample distribution area. This is also the reason why this collapsed GAN achieves a convergence rate as high as 97%.

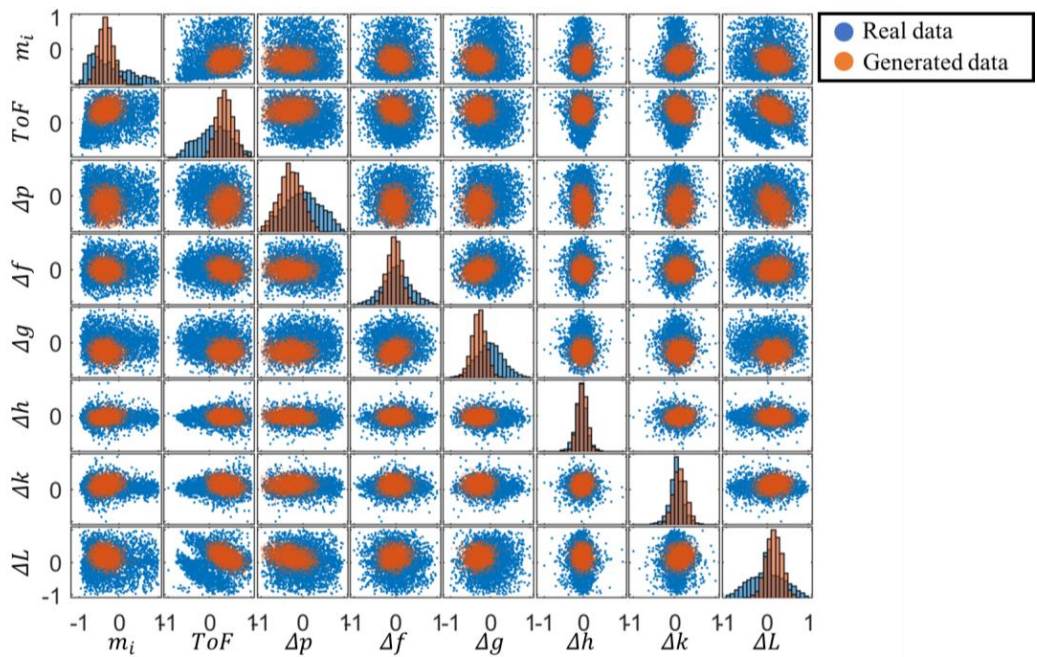

**Fig. 7  Distribution of generated samples from the mode collapsed GAN.**

Fig. 8 uses a box and whisker plot to illustrate detailed distributions of the real LT features and generated LT features. The data have been pre-processed and scaled between -1 and 1. Each box is drawn from the first quartile to the third quartile of the corresponding feature, the median value is marked by the target sign. The data points at the tail of whiskers are considered as outliers. It can be seen that the features generated by the GAN are closer to the real data statistics. For the collapsed GAN, we observed more outlies and the boxes and median points are largely biased from the real data boxes and medians.

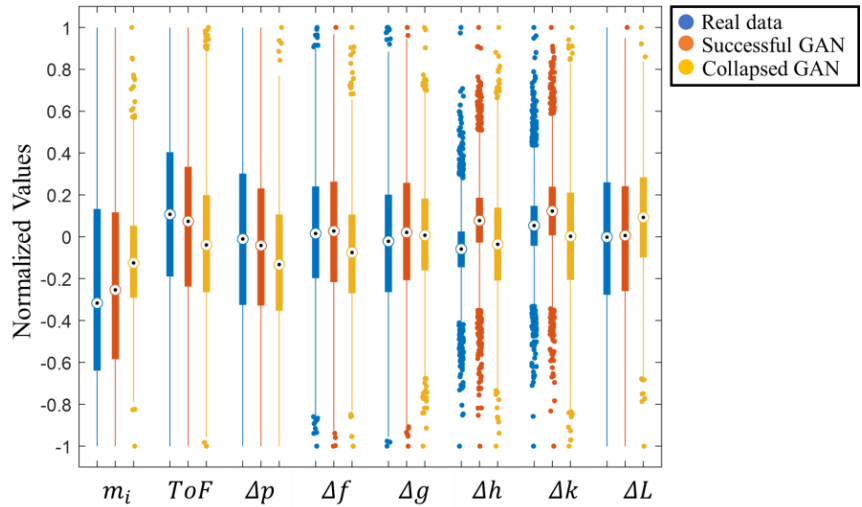

**Fig. 8  Box and whisker plots of real LT features, generated features from the successful GAN and collapsed GAN**

## V.  Conclusion

A GAN-based approach is proposed to boost the low thrust data generation efficiency for training deep learning models in the field of low thrust trajectory optimization. The valid GAN structure is identified, and the convergence rate of the GAN-generated low thrust transfer is improved to 84.3%. The GAN is tested in the low thrust NEA prospecting mission scenario, but the method may be mitigated to other contexts such as landing trajectory optimization.

The proposed technique is particularly useful when mission designers need to train multiple DNN models to deal with different mission scenarios and spacecraft configurations. In that case, multiple LT training databases need to be generated and the GAN can greatly reduce the computation cost.

Due to the complexity of GAN structure, much effort has been made to identify a valid GAN structure in the large hyperparameter search space. The structure of the successful GAN identified in this work is a good starting point to optimize GANs for data generation for other mission scenarios.

The GAN proposed in this work can be further improved in many aspects. First, the network can be further optimized by using other types of layers, such as convolutional layers, rather than fully connected layers. Second, the current GAN may be further improved by optimizing more hyperparameters.

## Acknowledgments

This research includes computations using the computational cluster Katana supported by Research Technology Services at UNSW Sydney and computation resources from the National Computational Infrastructure (project im32), which is supported by the National Computational Merit Allocation Scheme & UNSW High Performance Computing Resource Allocation Scheme. Financial support for this research was provided by the UIPA scholarship (RSRE7063, RSRE7061) from University of New South Wales (UNSW) and Commonwealth Scientific and Industrial Research Organisation (CSIRO) top-up scholarship.

## References

[1]  L. Cheng, Z. Wang, F. Jiang, and J. Li, "Fast Generation of Optimal Asteroid Landing Trajectories Using Deep Neural Networks," *IEEE Transactions on Aerospace and Electronic Systems,* vol. 56, no. 4, pp. 2642-2655, 2020, doi: 10.1109/taes.2019.2952700.

[2]	H. Yang, X. Bai, and H. Baoyin, "Rapid Trajectory Planning for Asteroid Landing with Thrust Magnitude Constraint," *Journal of Guidance, Control, and Dynamics,* vol. 40, no. 10, pp. 2713-2720, 2017, doi: 10.2514/1.G002346.
[3]	C. Sánchez-Sánchez and D. Izzo, "Real-Time Optimal Control via Deep Neural Networks: Study on Landing Problems," *Journal of Guidance, Control, and Dynamics,* vol. 41, no. 5, pp. 1122-1135, 2018, doi: 10.2514/1.G002357.
[4]	D. Izzo and E. Öztürk, "Real-Time Guidance for Low-Thrust Transfers Using Deep Neural Networks," *Journal of Guidance, Control, and Dynamics,* vol. 44, no. 2, pp. 315-327, 2021, doi: 10.2514/1.G005254.
[5]	R. Xie and A. G. Dempster, "An on-line deep learning framework for low-thrust trajectory optimisation," *Aerospace Science and Technology,* vol. 118, 2021, doi: 10.1016/j.ast.2021.107002.
[6]	H. Li, S. Chen, D. Izzo, and H. Baoyin, "Deep networks as approximators of optimal low-thrust and multi-impulse cost in multitarget missions," *Acta Astronautica,* vol. 166, pp. 469-481, 2020, doi: 10.1016/j.actaastro.2019.09.023.
[7]	Y. Song and S. Gong, "Solar-sail deep space trajectory optimization using successive convex programming," *Astrophysics and Space Science,* vol. 364, no. 7, 2019, doi: 10.1007/s10509-019-3597-x.
[8]	A. Zavoli and L. Federici, "Reinforcement Learning for Robust Trajectory Design of Interplanetary Missions," *Journal of Guidance, Control, and Dynamics,* vol. 44, no. 8, pp. 1440-1453, 2021, doi: 10.2514/1.G005794.
[9]	R. Xie and A. G. Dempster, "A Survey of Near-Earth Asteroids for Low-Thrust Round-Trip Missions," presented at the 2022 IEEE Aerospace Conference (AERO), 2022.
[10]	Y. Song and S. P. Gong, "Solar-sail trajectory design for multiple near-Earth asteroid exploration based on deep neural networks," (in English), *Aerospace Science and Technology,* vol. 91, pp. 28-40, Aug 2019, doi: 10.1016/j.ast.2019.04.056.
[11]	N. Qi, Z. Fan, M. Huo, D. Du, and C. Zhao, "Fast Trajectory Generation and Asteroid Sequence Selection in Multispacecraft for Multiasteroid Exploration," *IEEE Transactions on Cybernetics,* pp. 1-12, 2021-01-01 2021, doi: 10.1109/tcyb.2020.3040799.
[12]	H. Li and H. Baoyin, "Sequence optimization for multiple asteroids rendezvous via cluster analysis and probability-based beam search," *Science China Technological Sciences,* vol. 64, no. 1, pp. 122-130, 2021-01-01 2021, doi: 10.1007/s11431-020-1560-9.
[13]	G. Viavattene and M. Ceriotti, "Artificial Neural Networks for Multiple NEA Rendezvous Missions with Continuous Thrust," *Journal of Spacecraft and Rockets,* pp. 1-13, 2021, doi: 10.2514/1.A34799.
[14]	D. Izzo, C. I. Sprague, and D. V. Tailor, "Machine learning and evolutionary techniques in interplanetary trajectory design," in *Modeling and Optimization in Space Engineering*: Springer, 2019, pp. 191-210.
[15]	A. Shirazi, J. Ceberio, and J. A. Lozano, "Spacecraft trajectory optimization: A review of models, objectives, approaches and solutions," *Progress in Aerospace Sciences,* vol. 102, pp. 76-98, 2018, doi: 10.1016/j.paerosci.2018.07.007.
[16]	D. Hennes, D. Izzo, and D. Landau, "Fast approximators for optimal low-thrust hops between main belt asteroids," in *2016 IEEE Symposium Series on Computational Intelligence (SSCI)*, 6-9 Dec. 2016 2016, pp. 1-7, doi: 10.1109/SSCI.2016.7850107.
[17]	J. Sims, P. Finlayson, E. Rinderle, M. Vavrina, and T. Kowalkowski, "Implementation of a Low-Thrust Trajectory Optimization Algorithm for Preliminary Design," presented at the AIAA/AAS Astrodynamics Specialist Conference and Exhibit, 2006.
[18]	R. Xie and A. G. Dempster, "Feasible Low-thrust Trajectory Identification via a Deep Neural Network Classifier," *arXiv preprint arXiv:2202.04962,* 2022.